\documentclass[10pt,twocolumn,letterpaper]{article}

\usepackage{cvpr}              

\usepackage[dvipsnames]{xcolor}

\definecolor{cvprblue}{rgb}{0.21,0.49,0.74}
\usepackage[pagebackref,breaklinks,colorlinks,citecolor=cvprblue]{hyperref}

\usepackage{booktabs}
\usepackage{colortbl}
\usepackage{url}
\usepackage{wrapfig}
\usepackage{blindtext}

 \usepackage{multirow}
\usepackage[utf8]{inputenc} 
\usepackage[T1]{fontenc}    
\usepackage{amsfonts}       
\usepackage{nicefrac}       
\usepackage{microtype}      
\usepackage{xcolor}         

\def\paperID{222} 
\def\confName{3DV\xspace}
\def\confYear{2024\xspace}

\title{NCRF: Neural Contact Radiance Fields for Free-Viewpoint Rendering of Hand-Object Interaction}

\author{
Zhongqun Zhang\textsuperscript{1,2}\protect\thanks{} \and Jifei Song\textsuperscript{2}\protect\thanks{} \and Eduardo P\'{e}rez-Pellitero\textsuperscript{2} \and Yiren Zhou\textsuperscript{2} \and 
Hyung Jin Chang\textsuperscript{1} \and Ale\v{s} Leonardis\textsuperscript{1,2} \and
\textsuperscript{1}University of Birmingham, UK \quad  
\textsuperscript{2}Huawei, Noah’s Ark Lab \and 
{\tt\small zxz064@student.bham.ac.uk},
{\tt\small \{jifeisong,e.perez.pellitero,zhouyiren\}@huawei.com}\\
{\tt\small \{h.j.chang,a.leonardis\}@bham.ac.uk}
}

\begin{document}
\twocolumn[{
\maketitle
\begin{center}
    \captionsetup{type=figure}
    \includegraphics[width=\linewidth]{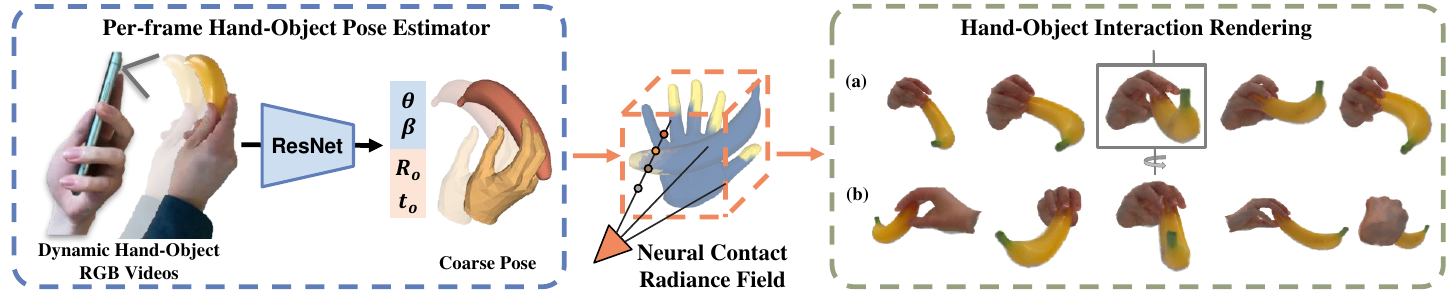}
    \caption{Given RGB videos of dynamic hand-object interaction and coarse hand-object poses generated by a pose estimator, NCRF models (a) hand-object movement with photo-realistic appearance, which can be used for  (b) free-viewpoint rendering for any frame (e.g. frame highlighted in the gray square) in a sequence.}
    \label{fig:highlevel_framework}
\end{center}
}]
{
  \renewcommand{\thefootnote}%
    {\fnsymbol{footnote}}
  \footnotetext[1]{This work was developed while interning at Huawei.}
  \footnotetext[2]{Corresponding author.}
}
\begin{abstract}
Modeling hand-object interactions is a fundamentally challenging task in 3D computer vision. Despite remarkable progress that has been achieved in this field, existing methods still fail to synthesize the hand-object interaction photo-realistically, suffering from degraded rendering quality caused by the heavy mutual occlusions between the hand and the object, and inaccurate hand-object pose estimation. To tackle these challenges, we present a novel free-viewpoint rendering framework, Neural Contact Radiance Field (NCRF), to reconstruct hand-object interactions from a sparse set of videos. In particular, the proposed NCRF framework consists of two key components: (a) A contact optimization field that predicts an accurate contact field from 3D query points for achieving desirable contact between the hand and the object. (b) A hand-object neural radiance field to learn an implicit hand-object representation in a static canonical space, in concert with the specifically designed hand-object motion field to produce observation-to-canonical correspondences. We jointly learn these key components where they mutually help and regularize each other with visual and geometric constraints, producing a high-quality hand-object reconstruction that achieves photo-realistic novel view synthesis. Extensive experiments on HO3D and DexYCB datasets show that our approach outperforms the current state-of-the-art in terms of both rendering quality and pose estimation accuracy. 
\end{abstract}
\section{Introduction}

Understanding and modeling hand-object interactions, as well as the reconstruction and manipulation, play an important role in interpreting human activities and behaviors~\cite{hasson2019learning,hasson2020leveraging,hasson20_handobjectconsist,cao2020reconstructing,doosti2020hope,tekin2019h+,Kwon_2021_ICCV,jiang2021hand,yang2021cpf,liu2021semi}. It is of particular interest for various of promising applications in human-computer interaction~\cite{ueda2003hand, sridhar2015investigating}, AR/VR~\cite{holl2018efficient, qian2022arnnotate}, and imitation learning in robotics~\cite{qin2022dexmv, garcia2020physics}. Previous works typically formulate this task as a joint hand and object pose estimation problem~\cite{hasson2020leveraging, doosti2020hope, li2021artiboost} and rely on parametric hand-object models such as MANO~\cite{romero2017embodied} and YCB~\cite{calli2015ycb} to estimate the hand motion transformation. Even though hand parametric models like MANO can bring strong prior knowledge of the shape information, existing methods struggle to recover the details and are limited to a lower resolution. One of the latest approaches, NIMBLE~\cite{li2022nimble}, further shows improved skin details over MANO but requires expensive scan data and annotation to generate a personalized and photo-realistic appearance. 

The direction of neural rendering sheds light on the photo-realistic novel view synthesis. Neural Radiance Field (NeRF)~\cite{mildenhall2020nerf} proposed to learn implicit neural representations, with the integration of position encoding and volumetric rendering, showing huge potential for high-quality novel-view synthesis and fine-grained detail reconstruction. NeRF has later been adapted to represent dynamic human body~\cite{peng2021animatable,peng2021neural,weng2022humannerf,jiang2022neuman,su2021nerf,te2022neural}, achieving compelling results on modeling a clothed human. Similarly, leveraging head parametric models like FLAME, 3D head avatar creation approaches~\cite{hong2022headnerf, athar2022rignerf, xu2022manvatar} built on NeRF also show high-quality facial appearance and geometry. In terms of hand modeling, LISA~\cite{corona2022lisa} is the first neural rendering model that can capture the accurate shape and appearance of human hands from multi-view images. The hand reconstruction quality has been further improved by the following work~\cite{chen2022hand}. However, how the hand is interacting with the outside world, specifically hand-object interaction is not yet considered in the light of neural rendering. Existing dynamic NeRF methods~\cite{weng2022humannerf, peng2021animatable} also fail to apply to hand-object interaction, due to the complex motion involved in human hand grasping objects, and heavy mutual occlusions frequently happening between hand and objects. Moreover, the visual and geometric prior of contact information is of critical importance but is unfortunately ignored in the previous neural hand approaches. 

To address the aforementioned challenges, we propose Neural Contact Radiance Field (NCRF), where we focus on the problem of free-viewpoints rendering for hand-object interactions using RGB videos, as illustrated in Fig.~\ref{fig:highlevel_framework}. Our method can work on multi-view video captured by hardware-synchronized industrial cameras in the lab setting, like DexYCB Dataset~\cite{chao2021dexycb}, and more importantly, can also work with monocular video taken from mobile phones in a casual way, making our pipeline widely applicable to hand-held mobile devices. To model and reconstruct the realistic details of the hand-object interaction, we build a novel dynamic hand-object neural radiance field that learns an implicit neural representation of hand-object interactions in canonical space, wherein a specifically designed hand-object motion field is incorporated to build observation-to-canonical correspondence, combining both skeletal and non-rigid motion for the hand, as well as the rigid motion transformation for the object. In addition, to further improve the hand-object pose estimation, we design a novel contact optimization field that refines the initial coarse hand-object poses from an off-the-shelf image-based pose estimator~\cite{hasson2020leveraging}, where we first estimate a contact field between hand and object by introducing an attention mechanism to model the interaction cues of 3D hand-object query points, and then constrain the hand-object pose with the desired contact prior. Lastly, to solve the challenges introduced by heavy mutual occlusion when hands often inevitably intersect with the object, we introduce a mesh-guided ray sampling strategy to mitigate the blurriness caused by hand-object intersection. We jointly optimize the contact optimization field and the neural radiance field using contact and photo-metric loss and observed that both hand-object interaction modeling and hand-object pose estimation can benefit each other, and achieve high-quality hand-object reconstruction performance.  

To the best of our knowledge, we are the first to propose a free-viewpoint rendering system based on NeRF for hand-object interactions. We evaluate our approach on the HO3D~\cite{hampali2020honnotate} and DexYCB~\cite{chao2021dexycb} datasets that capture hand-object interactions in complex motions. Across all video sequences, our approach exhibits state-of-the-art performances on free-viewpoint rendering and shows significant improvements in hand-object pose optimization. In summary, our contributions are three-fold:

\begin{itemize}
    \setlength{\itemsep}{0pt}%
    \setlength{\parskip}{0pt}%
    
    \item We propose a novel dynamic hand-object neural radiance field capable of modeling complex hand-object interactions with mutual occlusions between hand and object, achieving high-quality hand-object reconstructions and photo-realistic novel view rendering.

    \item To further improve the modeling of hand-object interaction, we propose a novel attention-based network for estimating the hand-object contact field and optimizing both hand and object poses, and the refined pose will further benefit the hand-object neural rendering under joint learning. 
    
    \item Extensive experiment results on HO3D and DexYCB show that our method outperforms the state-of-the-art both quantitatively and qualitatively, in terms of both hand-object rendering quality and hand-object pose estimation accuracy. 
\end{itemize}
\section{Related work}
Our work tackles the problem of hand-object interaction from a sparse set of videos. We first give the literature review on modeling hand-object interaction. Then we review the human-centric neural rendering. To this end, we briefly cover the methods using the contact information for hand-object pose refinement. 

\textbf{Hand-object interaction.}
modeling hand-object interaction is of promising application value yet challenging to solve due to severe occlusion between hand and object. Earlier work by Hasson~\etal~\cite{hasson2019learning} gave the first attempt by jointly regressing MANO hand parameters~\cite{romero2017embodied} and reconstructing object mesh with physical constraints, showing improved grasping quality. Most works~\cite{liu2021semi,hasson2020leveraging,hampali2022keypoint,li2021artiboost, cao2020reconstructing, yang2021cpf,grady2021contactopt, zhou2022toch, tse2022s, yu2022uv} assume the object models is known and estimate the 6D object pose instead. Leveraging the spatial-temporal consistency, Liu~\etal~\cite{liu2021semi} boosts the estimation performance through contextual reasoning to extract interaction cues in time. Other works focus on inferring the shape and pose of the hand-held object agnostically by lifting the object template constraints~\cite{hasson2020leveraging,tse2022collaborative,ye2022s,chen2022alignsdf}. For example, Chen~\etal~\cite{chen2022alignsdf} reconstructs hand-object shapes by combining SDFs and parametric representation and demonstrates the aligned SDF model reconstructs better shape details. Though existing approaches have shown improved hand grasping and object manipulation, the hand-object interaction modeling results lack the desired photo-realistic appearance information. In contrast, our work can recover the high-quality appearance as well as the detailed shape information.  

\textbf{Human-centric neural radiance field.}
Neural Radiance Field (NeRF)~\cite{mildenhall2020nerf} is capable of synthesizing photo-realistic novel views of a static scene from multi-view images by learning an implicit representation. Further explorations~\cite{weng2022humannerf, peng2021neural, peng2021animatable, su2021nerf,jiang2022neuman,te2022neural} adapt NeRF to deformable humans. NeuralBody~\cite{peng2021neural} proposes to integrate structured latent code diffused from SMPL meshes into the neural radiance field for dynamic human modeling. Furthermore, Peng~\etal~\cite{peng2021animatable} and Weng~\etal~\cite{weng2022humannerf} learn implicit body representation in a canonical T-pose space and learned blend weights are applied with 3D human skeletons to generate observation-to-canonical correspondences. To improve the generalization ability of neural human reconstruction, KeypointNeRF \cite{mihajlovic2022keypointnerf} takes use of relative 3D spatial information from sparse key points. Specific to human hand modeling, LISA~\cite{corona2022lisa} chooses the skeleton model parameterized by MANO and predicts the color and density from the learned implicit function. Chen~\etal~\cite{chen2022hand} extent MANO to MANO-HD and generate high-fidelity appearance from the neural radiance field learned from monocular videos. However, existing human-centric neural rendering methods ignore the hand-object interaction, leaving the challenging task of modeling complex interactions between hand and object within the neural radiance field unresolved. 

\textbf{Contact optimization.}
Contact optimization is one important auxiliary task for hand-object interaction. ContactOpt \cite{grady2021contactopt} has shown that by estimating contact regions on the object, hand pose estimation can be further improved in hand-object tracking. S$2$Contact~\cite{tse2022s} leverages the annotated contact map together with free-generated pseudo contact information under the semi-supervised learning framework, demonstrating better generalization ability across datasets. Recently, TOCH~\cite{zhou2022toch} is also built on contact optimization and further introduces a temporal denoising auto-encoder to generate plausible grasping sequences. In this work, we propose a novel attention-based mechanism to encode the contact information and incorporate contact optimization in learning a more accurate motion transformation for both hand and object, which will be beneficial to the implicit neural representation learning for hand-object interaction.  
\section{Method}
\begin{figure*}[htbp]
\centering
\includegraphics[width=1.0\linewidth]{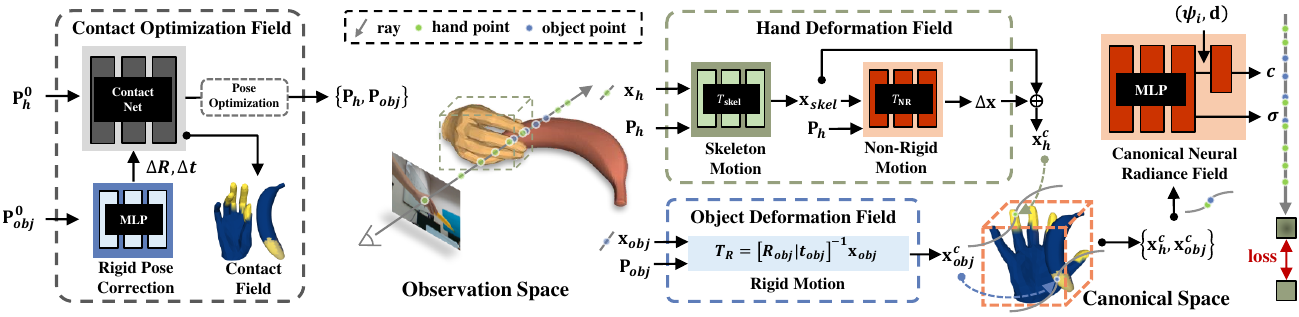}
\caption{
NCRF pipeline. We propose a novel hand-object neural radiance field to model the hand-object interaction. Our framework is composed of 1) a contact optimization field, leveraging contact prior in order to refine the hand-object pose, 2) a hand deformation field to deform hand points from observation to canonical space, considering both the skeletal and non-rigid motion, 3) object deformation field, which transforms the object into canonical space using refined rigid object pose, and 4) the canonical neural radiance field to build canonical volume for hand-object interaction and predict color and density. }
\label{fig:Pipeline}
\end{figure*}
Given a sparse set of videos of hand-object interaction, our task is to generate a free-viewpoint video of the scene of hand-object interaction and optimize the poses of hand and object simultaneously. Without loss of generality, for each frame, we extract the foreground hand-object mask and apply it to filter out the background image pixels. We also initialize per-frame hand-object pose and mesh using an image-based off-the-shelf pose and shape estimator~\cite{hasson2020leveraging}.

The overview of our approach is shown in Fig.~\ref{fig:Pipeline}, where we propose a novel hand-object neural radiance field to model the interaction between the hand and objects. We first propose the hand-object neural radiance field to model the hand-object interaction (Sec.~\ref{sec:hand-object nerf}). Moreover, a mesh-guided sampling method is used to separate the hand and object 3D query points based on the hand-object poses and mesh. To this end, we discuss how to estimate the contact field to refine the hand-object poses with a differentiable contact optimizer (Sec.~\ref{sec:contact optimization}), which will be jointly learned with the hand-object neural radiance field. 

\subsection{Hand-Object Neural Radiance Field}
\label{sec:hand-object nerf}

NeRF~\cite{mildenhall2020nerf} models the continuous radiance field of a static scene by learning an implicit neural function. As shown in Fig.~\ref{fig:Pipeline}, we further extend the neural radiance field to model the dynamic hand-object interaction. Specifically, for a given 3D point sampled from the scene of hand-object interaction, and a viewing direction as the input, our hand-object neural radiance field will be able to render the color and reconstruct the density. We first define the canonical volume, $F_c$, for the hand-object interaction: 

\begin{equation}
F_c:(\mathbf{x}_c, \mathbf{d}, \boldsymbol{\psi}_i) \mapsto(\mathbf{c}, \sigma),
\end{equation}
where $\mathbf{x}_c$ is the sampled 3D points from either hand or object along the ray direction in the canonical space. For each 3D point, it takes a spatial position $\mathbf{x}$ and viewing direction $\mathbf{d}$ as input to a neural network and outputs color $\mathbf{c}$ and volume density $\sigma$. Here we also include the latent appearance code for the $i$-th frame, $\boldsymbol{\psi}_i$, following NeRFies~\cite{park2021nerfies}. The function $F_c$ is modeled by a multi-layer perceptron (MLP) network, which is trained from a set of RGB images of a hand-object interaction scene captured with different poses.

We then define a hand-object deformation field $T$ to build the corresponding mapping for the 3D points between observation and canonical space on top of the hand-object pose. Thus, we can obtain the observation volume $F_o$ by warping the canonical volume $F_c$ with deformation field $T$:

\begin{equation}
F_o(\mathbf{x}, \mathbf{d}, \boldsymbol{\psi}_i)=F_c(T(\mathbf{x}, \mathbf{P}), \mathbf{d}, \boldsymbol{\psi}_i),
\end{equation}
where predicting the color $\mathbf{c}$ and density $\sigma$ for sampled 3D hand-object points $\mathbf{x}$ in observation space using observation volume $F_o$, is then converted to obtain the color and density of the corresponding point in canonical volume $F_c$, and $T$ deforms 3D points from observation to canonical space, guided by the hand-object pose $\mathbf{P}$.

\textbf{Hand-object deformation field.}
The motion transformation in the hand-object interaction is composed of hand skeleton motion, hand non-rigid motion, and object rigid motion. Therefore, we decompose the complex hand-object deformation field into the following parts:
\begin{equation}
T(\mathbf{x}, \mathbf{P})=
    \begin{cases}
        T_{\text{skel}}(\mathbf{x}, \mathbf{P}_h) + T_{\text{NR}}(\mathbf{x}, \mathbf{P}_{h}), \text{if}~\mathbf{x} \in \mathbf{X}_h  \\
        T_{\text{R}}(\mathbf{x}, \mathbf{P}_{obj}), \text{if}~\mathbf{x} \in \mathbf{X}_{obj}
    \end{cases}   
\end{equation}
where $T_{\text{skel}}$ provides the hand skeleton deformation, driven by refined hand pose $\mathbf{P}_h$, while $T_{NR}$ describes the non-rigid movement for hand points $\mathbf{X}_h$. $T_{R}$ provides rigid object deformation for object points $\mathbf{X}_{obj}$, driven by refined 6D object pose $\mathbf{P}$. Accurately separating 3D points into hand and object points is important, as the hand-object deformation falls into different motion fields of either hand or object. We, therefore, proposed a mesh-guided ray sampling strategy to indicate whether sampled points belong to hands or objects.

In particular, we first construct the hand deformation field based on the 3D hand skeleton. To map hand points from observation to canonical space, we compute the skeleton transformation using linear blend skinning algorithm~\cite{lewis2000pose}:
\begin{equation}
T_{\text{skel}}(\mathbf{x}, \mathbf{P}_h)=\sum_{i=1}^K w_o^i(\mathbf{x})(R_i\mathbf{x}+t_i)
\end{equation}
where $w_o^i$ is the blend weight for the $i$-th bone in observation space. $R_i$ and $t_i$ are rotation and translation for the $i$-th bone, respectively, which can be explicitly computed from hand pose $\mathbf{P}_h$ \cite{romero2017embodied}. Following the HumanNeRF~\cite{weng2022humannerf}, we use a CNN to learn blend weights $\mathbf{W}_c$ in canonical space, from which we can derive $w_o^i$ (see supplementary).

Furthermore, we build an additional non-rigid deformation field via MLP to model free-form shape deformation, \eg, the deformation of fingertips due to pressure. It learns an offset $\Delta \mathbf{x}$ for $\mathbf{x}_{\text{skel}}$ in contact region conditioned on joint rotation.

Opposite to the hand, we build the object's deformation field as a rigid motion field, $T_{\text{R}}$. For each sampled point inside the object, $\mathbf{x} \in \mathbf{X}_{obj}$, we convert it to object canonical space with the inverse transformation of object pose:
\begin{equation}
T_{\text {R}}=\left[R_{obj} \mid t_{obj}\right]^{-1}\mathbf{x},
\end{equation}
where $R_{obj}$ and $t_{obj}$ are the rotation and translation in the 6D object pose $\mathbf{P}_{obj}$, respectively.

\textbf{Mesh-guided ray sampling.}
As mentioned above, since the hand and object follow different motion transformations from observation to canonical space, we need to separate hand and object points in the observation space. However, if we simply follow~\cite{weng2022humannerf} to leverage the 3D bounding box or 2D segmentation map for separating and sampling points, the hand points will inevitably intersect with the object points due to ambiguity and inaccurate separation. To alleviate this issue,  we design a mesh-guided ray sampling scheme, so that hand and object point space will not penetrate each other. Considering the rigid object has a pre-defined shape prior, we could directly constrain sampled points inside the object mesh as the object point set, $\mathbf{X}_{obj}$:  
\begin{equation} 
\mathbf{X}_{obj}=\{\mathbf{x} \in \mathcal{HO} | \Psi(\mathbf{x}, \mathcal{M}_{obj}) <= 0\}
\end{equation}
where $\mathcal{M}_{obj}$ is object mesh, and $\mathcal{HO}$ denotes the set of all sampling points in the hand-object bounding box. $\Psi$ indicates whether a point is located outside or inside the given 3D shape. The rest of the points will be viewed as belonging to the hand point set, $\mathbf{X}_{h}$, and will be further constrained by the foreground information reflected from $\mathbf{W_c}$ to focus on the hand part in the foreground.
\begin{figure*}[htbp]
\centering
\includegraphics[width=1.0\linewidth]{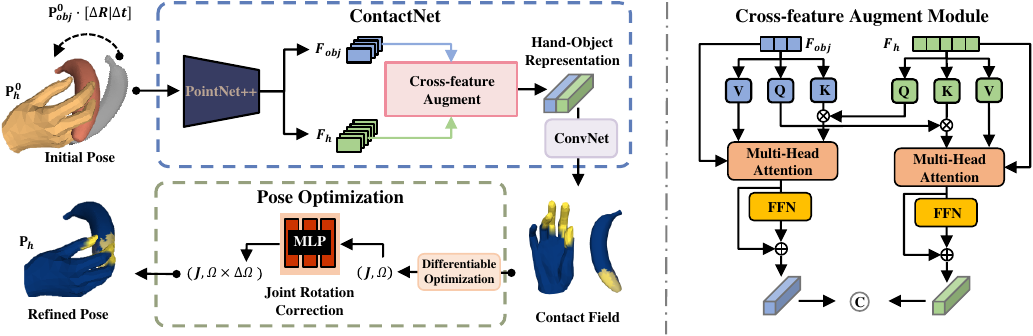}
\caption{
Structure of contact optimization field. Given initial hand-object poses $\boldsymbol{P}_h^0$, $\boldsymbol{P}_{obj}^0$, we first refine object pose by a residual $\Delta R, \Delta t$ learned from Rigid Pose Correction module (not drawn). Our ContactNet consists of 1) a PointNet++ backbone to extract features for both hand and object, 2) the attention-based cross-feature augment to obtain hand-object interaction representations, and 3) the ConvNet to regress the contact field. The Pose Optimization consists of 1) the Differential Optimization (DiffOpt) module, iteratively updating hand joint and rotation $(J, \Omega)$ conditional on the contact field, 2) the Joint Rotation Correction module to learn a rotation residual $\Delta \Omega$, with which the final refined hand pose is obtained. 
}
\label{fig:ContactNet}
\end{figure*}

\subsection{Contact Optimization Field}
\label{sec:contact optimization}
The hand-object pose is of high importance to the hand-object deformation field. Nevertheless, the estimation from the off-the-shelf hand-object estimator is not accurate and thus cannot meet our needs for learning the high-quality dynamic hand-object neural radiance field. Therefore, we leverage contact prior in order to obtain an accurate estimation for the hand-object pose. As shown in Fig.~\ref{fig:ContactNet}, given a hand mesh $\mathcal{M}_{h}$ and object mesh $\mathcal{M}_{obj}$ with initial coarse pose $\mathbf{P}^0=\left\{\mathbf{P}_h^0, \mathbf{P}_{obj}^0\right\}$, we propose the contact optimization field, which first learns to estimate the contact field $\mathbf{C}$ between hand and object mesh, and further obtain the optimized the hand pose $\mathbf{P}_h$ and object pose $\mathbf{P}_{obj}$. 

\textbf{Rigid object pose correction.} Current contact optimizers, e.g. ContactOpt~\cite{grady2021contactopt} and TOCH~\cite{zhou2022toch}, have limited application scope as they only focus on refining hand pose while assuming the ground truth object pose is available or the estimated object pose is accurate. However, such an assumption is not true as either the ground truth object pose is very likely not annotated due to expensive labor work, or the estimated pose for the object could be inaccurate and need further refinement, especially when mutual occlusion is happening. To address this issue, we design a rigid object pose correction module to refine the initial object pose $\mathbf{P}_{obj}^0$:
\begin{equation}
\Delta R, \Delta t = \operatorname{MLP}(\mathbf{P}_{obj}^0),
\end{equation}
where $\Delta R$ is the residual update of rotation, represented by a 4-dimensional quaternion vector, and $\Delta t$ is the residual of translation. Both $\Delta R$ and $\Delta t$ will then be applied to refine the original object pose:

\begin{equation}
\mathbf{P}_{obj}=\mathbf{P}_{obj}^0 \cdot[\Delta R \mid \Delta t],
\end{equation}

Similar to DeepIM~\cite{li2018deepim}, our proposed module will iteratively refine the object pose by minimizing the photometric error between rendered image and the observed image. In other words, we jointly train this module with the hand-object neural render, where the photo-metric loss will in return help to refine the object pose in an end-to-end manner.

\textbf{ContactNet.} After we refine the object pose, we propose an attention-based network, ContactNet, to estimate the contact field between the hand-object and improve the hand pose estimation jointly, as shown in Fig.~\ref{fig:ContactNet}. The input for this module are coarse hand poses $\mathbf{P}_h^0$ and refined object pose $\mathbf{P}_{obj}$ and the output is the contact field: 
\begin{equation}
\mathbf{G} = \left\{\mathbf{G}_h \in \mathbb{R}^{N_h}, \mathbf{G}_{obj}\in \mathbb{R}^{N_{obj}}\right\},
\end{equation}
where $N_h$ and $N_{obj}$ are the number of 3D query hand-object points sampled from hand-object mesh surface, respectively. The contact value is normalized to the range [0, 1], indicating the probability of whether a query point has a corresponding contact point. Given the sampled hand-object query points, we use PointNet++~\cite{qi2017pointnet++} to extract hand-object features $F_h \in \mathbb{R}^{N_h \times 1024}$ and $F_{obj} \in \mathbb{R}^{N_{obj} \times 1024}$. Then we apply the cross-feature augment (CFA) module between two features and send the fused hand-object latent representation into a 1-D convolution network to regress the contact value. The structure of CFA module is shown on the right side of Fig.~\ref{fig:ContactNet}, which exploits the interaction between hand and object in the contact region and generates enhanced hand-object representations. 

\textbf{Hand pose optimization.}
Hand pose is then optimized in a differentiable way with obtained contact information. Our target is to find the hand pose that best complies with contact prior. Firstly, we apply the differentiable optimization (DiffOpt) module to get the contact field $\mathbf{{G}}_h$, $\mathbf{{G}}_{obj}$ based on the corrected hand pose~\cite{grady2021contactopt}. Secondly, we iteratively minimize the difference between the current contact field and the target contact field and update the hand pose parameters.

The above optimization ensures the grasping is physical-plausible by contact constraints, but there still exists slight misalignment between the optimized pose and the actual one reflected in the image. To bridge this gap, we further employ an MLP to learn a residual $\Delta\Omega$ for hand joint rotation by leveraging visual constraints. 
With this joint rotation correction module, the final refined hand pose can be obtained as:
\begin{equation}
\mathbf{P}_h=\left(J, \Delta{\Omega} \cdot \Omega, R_h, t_h \right).
\end{equation}
where $J$ are 3D joint locations, $\Omega$ are local joint rotations, and $R_h, t_h$ are the hand global rotation and translation, respectively, following the definition in MANO\cite{romero2017embodied}.

\subsection{Training}
The NCRF is trained with a photometric reconstruction loss for the estimated color $\mathbf{C}(\mathbf{r})$ of a ray $\mathbf{r}$ by rendering the hand-object radiance field using volume rendering, with regard to the ground truth color $\hat{\mathbf{C}}(\mathbf{r})$, including VGG-based LPIPS loss~\cite{zhang2018unreasonable} and MSE loss~\cite{mildenhall2020nerf}, with $\lambda_1=0.2$, under patch-based ray sampling following\cite{weng2022humannerf}:
\begin{equation}
\mathcal{L}_{\text{nerf}}=\mathcal{L}_\text{LPIPS} + \lambda_1\mathcal{L}_\text{MSE}.
\end{equation}

The ContactNet is pre-trained on the ContactPose dataset~\cite{brahmbhatt2020contactpose} with contact labels gathered from the thermal camera. 
The contact prediction from the ContactNet will then serve as pseudo label as the target contact map. Then We iteratively optimize the hand pose in the DiffOpt module by minimizing the contact optimization loss:
\begin{equation}
\mathcal{L}_\text{con}=\mid\mathbf{G}_{obj}-\mathbf{\hat{G}}_{obj}\mid + \lambda_2 \mid\mathbf{G}_h-\mathbf{\hat{G}}_h\mid,
\end{equation}
where $\mathbf{\hat{G}}_{obj}$ and $\mathbf{\hat{G}}_h$ are target contact maps. $\mathbf{G}_{obj}$ and $\mathbf{G}_h$ are contact maps interpreted from DiffOpt under estimated hand-object pose. We set $\lambda_2=3$. Besides, we incorporate a collision loss to avoid hand-object interpenetration:
\begin{equation}
\mathcal{L}_\text{pen}=\sum_{\boldsymbol{v} \in \mathcal{M}_{obj}} -\min({\text{SDF}}_h(\boldsymbol{v}),0),
\end{equation}
where $\text{SDF}_h(\cdot)$ is the signed distance field for the hand to check if any given vertex is inside the hand, and $\boldsymbol{v}$ represents vertex sampled from object mesh $\mathcal{M}_{obj}$. Finally, we obtain the overall objective for jointly optimizing neural rendering and contact optimization:
\begin{equation}
\mathcal{L} = \lambda_\text{nerf}\mathcal{L}_\text{nerf} + \lambda_\text{con}\mathcal{L}_\text{con} + \lambda_\text{pen}\mathcal{L}_\text{pen}.
\end{equation}
where $\lambda_\text{nerf}=1$, $\lambda_\text{con}=0.5$, and $\lambda_\text{pen}=0.5$ are the loss weights for the corresponding loss. 
\section{Experiments}
\begin{figure*}[htbp]
\centering
\includegraphics[width=1.0\linewidth]{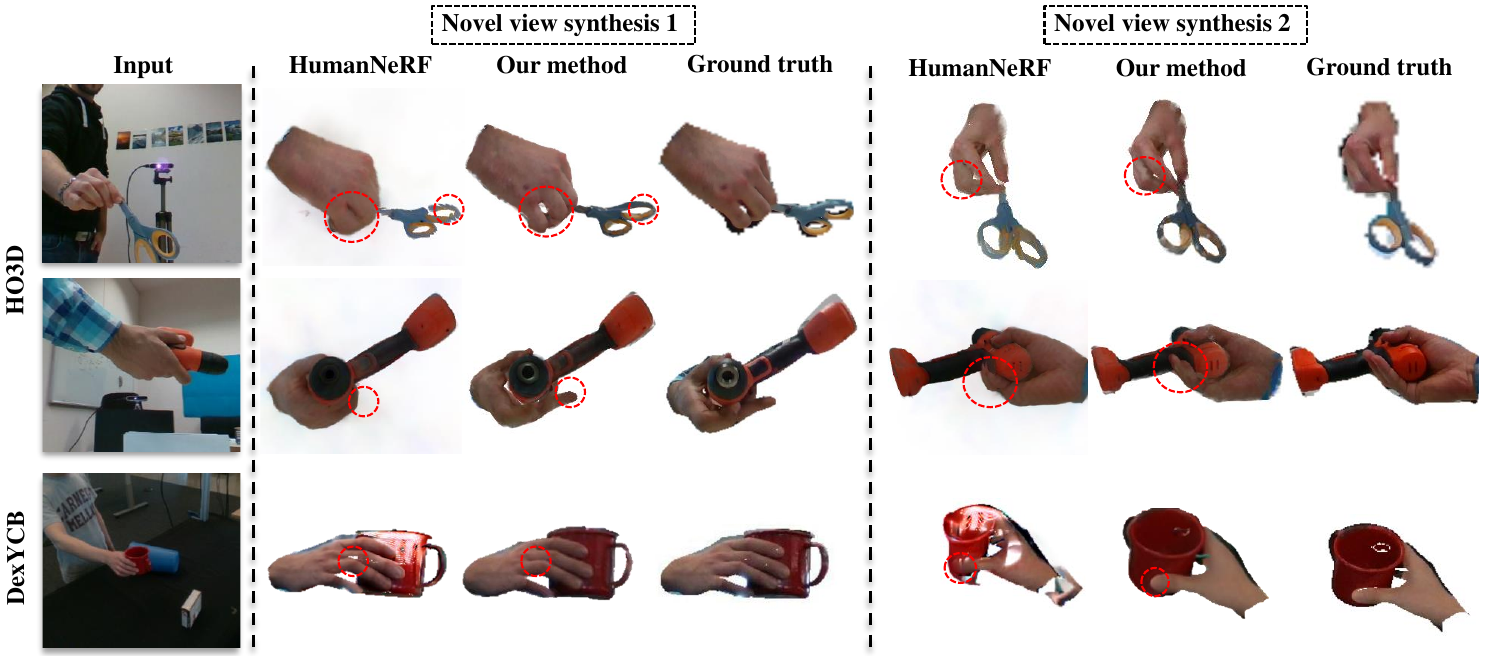}
\caption{
Qualitative comparison with HumanNeRF~\cite{weng2022humannerf} on the HO3D and DexYCB dataset.
}
\label{fig:Qulitative}
\end{figure*}
\begin{figure*}[htbp]
\centering
\includegraphics[width=1.0\linewidth]{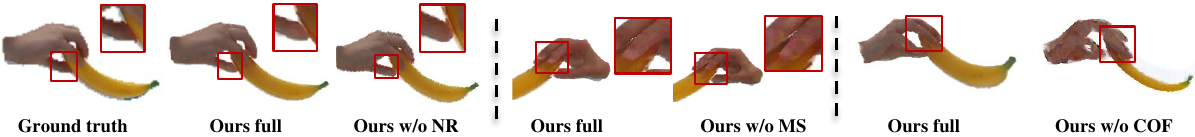}
\caption{
Ablation study on the hand-object neural radiance field consists of non-rigid motion (NR), mesh-guided sampling (MS) and contact optimization field (COF) modules. Zoom in for details.}
\label{fig:Ablation}
\end{figure*}

\subsection{Datasets and Metrics}

\textbf{Datasets.} 
We evaluate our method on HO3D~\cite{hampali2020honnotate} and DexYCB~\cite{chao2021dexycb} datasets, both of which capture dynamic hand-object interactions from multiple views and provide the annotations. Specifically, DexYCB is captured with 8 cameras, while HO3D is captured by 5 and we are using v3 where the camera poses are provided. To evaluate free-viewpoint rendering, we select multi-view videos with complex hand-object interaction yet enough appearance information. Following the protocol of ZJU-MoCap~\cite{peng2021neural}, we take three views for training and leave the remaining views for testing. More details can be found in the supplementary material.

\textbf{Metrics.}
Similar to preceding neural rendering approaches, we report PSNR, SSIM~\cite{wang2004image}, and LPIPS*~\cite{zhang2018unreasonable} (LPIPS* = LPIPS $\times 10^3$), as the metrics for evaluating rendering quality. Moreover, we report the Mean Per-Joint Position Error (MPJPE) and the Intersection Volume (IV) as the metrics of hand pose estimation and contact optimization, respectively, following ContactOpt~\cite{grady2021contactopt}. For object pose estimation, we report the metric of ADD-0.1D, which denotes the percentage of average object 3D vertices error within 10\% of object diameter.

\subsection{Comparison with State-of-the-art Methods}
\textbf{Baselines.} For free-viewpoint rendering, we compare our approach with a SOTA free-view synthesis method, HumanNeRF~\cite{weng2022humannerf}. In order to adapt it to our task, which is originally designed for neural body rendering, we first train the HumanNeRF based on the MANO for hand modeling and take our object branch for object modeling, and then register the coordinate systems in which two NeRF models align.  
For hand-object pose estimation, we compare our method with SOTA methods, ContactOpt~\cite{grady2021contactopt}, S$^2$Contact~\cite{tse2022s}, and TOCH~\cite{zhou2022toch}. All of their results are obtained using the officially released model and code and re-trained under our experiment setting for a fair comparison. 

\textbf{Comparison on rendering quality.} 
Given video frames for hand-object interaction, we synthesize novel views and make comparisons on rendering quality. From qualitative results shown in Fig.~\ref{fig:Qulitative}, we can observe that our visual quality is substantially better than HumanNeRF on both HO3D and DexYCB datasets. Although HumanNeRF also embeds a pose correction module to improve the alignment, our method can achieve much better reconstruction results, with the hand-object pose well refined by the designed pose correction module leveraging contact prior, as highlighted in the red circle. Quantitatively, as shown in Table~\ref{table:rendering_quality}, our method outperforms HumanNeRF~\cite{weng2022humannerf} on all the metrics with a large margin. The result again supports our method is able to achieve the best grasping rendering performance. 

\begin{table}[t]
  \caption{Quantitative results of free-viewpoint rendering on HO3D and DexYCB datasets. }
  \label{table:rendering_quality}
  \scriptsize
  \centering
  \setlength\tabcolsep{3pt}
  
  \resizebox{\columnwidth}{!}{
  \begin{tabular}{c |cc | cc | cc }
    \toprule
    \multirow{2}{*}{\textbf{Datasets}}
     & \multicolumn{2}{c|}{\textbf{PSNR}$\uparrow$}  & \multicolumn{2}{c|}{\textbf{SSIM}$\uparrow$}  & \multicolumn{2}{c}{\textbf{LPIPS*}$\downarrow$} \\
       & HumanNeRF\cite{weng2022humannerf} & Ours & HumanNeRF\cite{weng2022humannerf} & Ours & HumanNeRF\cite{weng2022humannerf} & Ours\\
    \midrule
     HO3D &26.06  &\textbf{29.74}    &0.9665  &\textbf{0.9758}   & 40.23 &\textbf{32.47} \\
     \midrule
     DexYCB &27.32  &\textbf{32.16}    &0.9719  &\textbf{0.9813 }  &27.43 &\textbf{20.37} \\
    \bottomrule
  \end{tabular}
  }
\end{table}
\begin{table}[t]
  \caption{Quantitative results of pose estimation on HO3D and DexYCB compared to state-of-the-art approaches. }
  \label{table:pose_optimization}
  \scriptsize
  \centering
  \setlength\tabcolsep{3pt}
  \begin{tabular}{l | l |c| c| c}
    \toprule
    \textbf{Datasets} &\textbf{Method}
    &\textbf{MPJPE (mm)}$\downarrow$  &\textbf{ADD-0.1D (\%)}$\uparrow$ &\textbf{IV ($\mathrm{cm}^3$)}$\downarrow$\\
    \midrule
    \multirow{5}{*}{HO3D~\cite{hampali2020honnotate}}
    &Hasson~\etal~\cite{hasson2020leveraging} & 11.4 & 74.5    & 9.3 \\
    &ContactOpt~\cite{grady2021contactopt} & 9.5 & -    & 8.1 \\
    &TOCH~\cite{zhou2022toch} & 9.3 & -  & 4.7 \\
    &S$^2$Contact~\cite{tse2022s} & \textbf{8.7} &81.4 & \textbf{3.5} \\
    &Ours & 8.9 &\textbf{89.8} & 3.7 \\
    \midrule
    \multirow{2}{*}{DexYCB~\cite{chao2021dexycb}}
    &S$^2$Contact~\cite{tse2022s} & 11.8 &70.5 & 10.5 \\
    &Ours &\textbf{10.2} &\textbf{83.2} & \textbf{9.3} \\
    \bottomrule
  \end{tabular}
\end{table}

\textbf{Comparison on pose estimation.} 
We further compare our method with the state-of-the-art method, S$^2$Contact, regarding the performance of hand-object pose estimation, and both methods use the output from the off-the-shelf pose estimator, Hasson~\etal~\cite{hasson2020leveraging}, as the initial pose. As shown in Table~\ref{table:pose_optimization}, we are able to achieve much lower hand pose error compared to S$^2$Contact on DexYCB dataset while on HO3D, achieving competitive performance on MPJPE and IV metric, and we improve 15.3\% and 8.4\% on ADD-0.1D metric over the initial pose from Hasson~\etal~\cite{hasson2020leveraging} and S$^2$Contact, respectively. We further include ContactOpt~\cite{grady2021contactopt} and TOCH~\cite{zhou2022toch} as the baselines on HO3D, and again our method achieves the best performance. Both ContactOpt and TOCH suffer the inaccurate object pose annotation, while our method is able to fully optimize the initial coarse hand-object pose, benefiting from both contact optimization and neural rendering.

\subsection{Ablation Studies}
We conduct the ablation study on HO3D dataset. We first explore the effect of different modules in the contact optimization field and hand-object neural radiance field respectively. After that, we analyze the benefit of joint learning contact optimization and neural rendering and the impact of different numbers of camera views.

\textbf{Ablation on contact optimization field.} We ablate our full model to study the effect of the contact optimization field, consisting of the cross-feature augment, rigid pose correction, and joint rotation correction module. For cross-feature augment, we compare the effect of different feature augment designs. As shown in Table~\ref{table:ab_pose_opt}, compared to our full model, the hand pose error (MPJPE) increases by 0.3mm and 0.2mm with enhanced hand feature only and the other way around, respectively, and adds up to 0.5mm after removing the whole module. In the ablated version without rigid pose correction, we also observe the hand pose error significantly increases by 2mm, showing that inaccurate object pose will hinder the contact optimization to get the correct hand pose. In the end, ablating the joint rotation correction will result in 0.1mm increment on MPJPE, and the IV error increases by 0.2 $\mathrm{cm}^3$. Therefore, the ablation study validates the effectiveness of the different modules of the contact optimization field. 

\begin{table}[t]
  \caption{Ablation study results of pose estimation.} 
  \label{table:ab_pose_opt}
  \scriptsize
  \centering
  \setlength\tabcolsep{3pt}
  \begin{tabular}{ l |c| c| c}
    \toprule
    \textbf{Model}
    &\textbf{MPJPE (mm)}$\downarrow$  &\textbf{ADD-0.1D (\%)}$\uparrow$ &\textbf{IV ($\mathrm{cm}^3$)}$\downarrow$\\
    \midrule
    hand-feature augment only & 9.1 & -    & 3.9 \\
    object-feature augment only & 9.2 & -  & 4.4 \\
    w/o cross-feature augment & 9.4 & -    & 5.2 \\
    \midrule
    w/o rigid pose correction & 10.9 &74.5 & 5.1 \\
    w/o joint rotation correction & 9.0 & - & 3.9\\
    w/o neural rendering &9.3 & 74.5 & 4.9\\
    \midrule
    full model & \textbf{8.9} &\textbf{89.8} &\textbf{3.7} \\
    \bottomrule
  \end{tabular}
\end{table}

\textbf{Ablation on hand-object neural radiance field.}
We then give the ablation study on the hand-object neural radiance field, where we investigate the effect of the proposed mesh-guided ray sampling and non-rigid deformation field. To validate the usefulness of mesh-guided sampling, we replace it with common 3D bounding box sampling. The first row of Table~\ref{table:ab_ho_nerf} shows that the superiority of our mesh-guided sampling, \eg, LPIPS* metric increases by 2.8 after replacement. From the visual comparison (Fig.~\ref{fig:Ablation}), we observe that the hand appearance will be contaminated by the colors blended from the object due to ambiguity and inaccurate separation without mesh-guided sampling, demonstrating the effective role of mesh-guided sampling in separating hand and object points. Furthermore, disabling non-rigid deformation field introduces a small increase in the LPIPS* metric by 2.15 (see the second row of Table~\ref{table:ab_ho_nerf}), thus reflecting the improvement brought by the non-rigid deformation field. Similarly, as shown in Fig.~\ref{fig:Ablation}, adding non-rigid motion can model free-form shape deformation, leading to more plausible grasping rendering. 

\textbf{Impact of joint learning and camera views.}
We study the effect of jointly learning contact optimization and neural rendering. As shown in the second last row in Table~\ref{table:ab_pose_opt}, the hand error increases by 0.4mm without the help of neural rendering. On the other hand, the performance of novel view synthesis significantly decreases without the help of contact optimization, as shown in the third row of Table~\ref{table:ab_ho_nerf}. We also compare our models trained with different numbers of camera views. The results in Table~\ref{table:ab_ho_nerf} show that overall the number of training views improves the performance of novel view synthesis. And it is worth noting that our model can also work effectively under the monocular video setting which is much more challenging.

\begin{table}[t]
  \caption{Ablation study results of novel view synthesis.} 
  \label{table:ab_ho_nerf}
  \scriptsize
  \centering
  \setlength\tabcolsep{3pt}
  \begin{tabular}{ l |c| c| c}
    \toprule
    \textbf{Model} &\textbf{PSNR}$\uparrow$  &\textbf{SSIM}$\uparrow$ &\textbf{LPIPS*}$\downarrow$\\
    \midrule
    w/o mesh-guided ray sampling & 28.10 &0.9693    & 35.33 \\
    w/o non-rigid motion & 28.81 & 0.9712    & 34.62 \\
    w/o pose optimization field & 25.71 & 0.9624    & 46.34\\
    \midrule
    with 2 camera views & 29.57 & 0.9732    & 33.01 \\
    with 1 camera view & 27.93 &0.9704    & 35.85 \\
    \midrule
    full model (3 camera views) & \textbf{29.74} &\textbf{0.9758} &\textbf{32.47} \\
    \bottomrule
  \end{tabular}
\end{table}
\section{Conclusion}
In this paper, we propose Neural Contact Radiance Field (NCRF), producing state-of-the-art performances for free-viewpoint renderings of hand-object interaction from a sparse set of videos. Our approach designs a dynamic hand-object neural radiance field capable of modeling challenging hand-object interaction, with complex hand-grasping motions and frequent mutual occlusions. The contact optimization field leverages the contact prior and refines the hand-object pose to comply with the contact constraints and the final image rendering. The refined pose will then drive the proposed hand-object deformation module with the help of mesh-guided sampling to deform rays into the canonical space and render the hand-object interaction photo-realistically. Extensive experiment results on various datasets demonstrate that our approach can achieve state-of-the-art hand-object reconstruction with photo-realistic free-view synthesis.

\textbf{Limitations.}
Our method has shown state-of-the-art performance in modeling hand-object interaction with photo-realistic appearance. However, our method still has a few limitations. 1) Our method will require the hand and object largely covered under the monocular video setting. 2) We use per-frame pose estimation, thus one following attempt will be applying temporal consistency. 3) We do not consider lighting conditions in this work. The next step will be modeling the environmental lighting conditions and enabling novel lighting synthesis. In summary, these limitations point to a range of exciting avenues for future work.
\section{Acknowledgements}
This work was supported by the Huawei London Research Center (2012 Laboratories), the Institute of Information \& communications Technology Planning \& Evaluation (IITP) grant funded by the Korean government(MSIT) (No.2022-0-00608, Artificial intelligence research about multi-modal interactions for empathetic conversations with humans).

\newpage

{\small
\bibliographystyle{ieeenat_fullname}
\bibliography{main}
}

\end{document}